# NLP-Driven Knowledge Extraction and Thematic Classification of Translated Ancient Indian Medical Texts

**Rajeevan M S**
Freelance Information Scientist,
Trivandrum, Kerala, India.
ORCID - 0009-0009-3320-3078
Mob: +91-7403663101
Email: msrajeevan96@gmail.com

**Dr. B. Mini Devi**
Assistant Professor
Department of Library and Information Science
University of Kerala, Palayam, Trivandrum, India.
ORCID - 0000-0003-2736-7999
Mob: +91-9446663070
Email: drminidevi1968@gmail.com

**Dr. V. S. Anoop**
Assistant Professor (Systems and Analytics)
Thiagarajar School of Management (Autonomous)
Madurai, India
ORCID - 0000-0001-6673-6932
Mob: 9747091417
Email: anoop@tsm.ac.in

**Dr. C. Mallikarjuna.**
Deputy Librarian
Knowledge Resource Centre (KRC)
Indian Institute of Technology Hyderabad (IITH)
Kandi, Sangareddy, Telangana, India.
ORCID- 0000-0003-3430-8903
Mob: +91-9494425022
Email: mallikarjuna.c@admin.iith.ac.in

**Abstract**

*Ancient Indian medical texts like Sushruta Samhita have extensive information on diseases, treatments, and surgical techniques. Yet, their ancient format and use of intricate vocabulary pose difficulties in accessibility and systematic ordering. The research here utilizes Natural Language Processing (NLP) methods like Named Entity Recognition (NER), BERTopic modeling, and Knowledge Graph development in Neo4j to extract, categorize, and visualize important concepts based on translated versions. Thematic classification with BERTopic allows for the identification of the underlying medical topics, whereas NER supports the structured entity recognition of diseases, treatments, researchers, and medicinal plants. Graph-based network analysis with Neo4j also allows for the semantic representation of relationship among extracted entities, supporting knowledge retrieval and digital preservation. The findings illustrate how graph databases, topic modeling, and entity recognition facilitate the computational organization of Ayurveda's historical medical wisdom, closing the gap between the conventional texts and contemporary data-driven inquiry. The suggested method promotes historical text analysis, medical informatics, and digital humanities to make ancient Indian medical wisdom more accessible and understandable.*

## 1. Introduction

Ancient Indian medical texts such as Sushruta Samhita are the foundation of Ayurveda, yet they contain some in-depth information on surgical procedures, pharmacology, and disease treatment. Still, their complex language patterns, chaotic organization, and capricious systematization have offered a momentous barrier to the touched and understand the information therein contained. Recent research on natural language processing (NLP) has made automatic extraction, classification, and representation of such complex information easy and feasible for integration into modern medical education (Bahad et al., 2024).

Some NLP methods according to machine learning(Jickson et al., 2023), like Named Entity Recognition (NER), subject-relevant classification, and knowledge graphs, enable programmatic identification within these historical documents of such key concepts as disease, treatment, and practitioner(Anoop et al., 2024). For instance, (Bahad et al., 2024) fine-tuned pre-trained NER models for Indian languages to address constraints in multilingual named entity recognition and expand the applicability of NLP techniques to various types of text.

Moreover, the application of NLP in the processing of various historical documents is automated upon segmentation, entity recognition, and topic modeling(Akhil et al., 2020). (Gurav, 2024) provides an overview of its application in electronic health records, giving attention to information extraction and how it enhances healthcare communication. Despite this broad feature in progress, it remains a largely occupational concept for Indian medical literature, with large vocabularic diversity, content structure that is somewhat odd, and multilingual texts. Sarella and (Sarella & Mangam, 2024) reviewed AI-aided NLP to facilitate health communication and showed how formal entity recognition improves the accuracy and efficiency of information extraction regarding disease and treatment information from historical backgrounds.

Recent studies resorted to deep learning and NLP in order to read older medical texts as a response to these issues. (Srivastava et al., 2024) devised a semi-supervised learning-based machine tool named MedPromptExtract-a combination of large language models, NLP, and prompt engineering-that transforms unstructured medical data into well-structured data, affirming the utility of NLP in putting together and interpreting complicated medical datasets. In continuation, (Mullick et al., 2023) also developed an intent classification and entity extraction model specialized for Indic languages-strongly reinforcing the need for domain-specific NLP approaches to medical and historical texts.

The way such medical texts are composed in ancient India, primarily the Sushruta Samhita, makes modern scientific study of them and access to knowledge therein a Herculean task. Range of NLP approaches for extracting, categorizing and providing primary medical knowledge are adopted in this work using Named Entity Recognition, BERTopic modeling and development of a knowledge graph in Neo4j, to facilitate structured and computational interpretation of Ayurvedic knowledge. The major contributions of this article are summarized as follows:

- Entity Recognition for Ayurvedic Knowledge Extraction: Utilizes NER to recognize diseases, treatments, medicinal plants, and scholars.
- Thematic Classification with Topic Modeling: Uses BERTopic to discover prominent medical themes.
- Knowledge Graph Construction: Creates a Neo4j-based knowledge graph to capture relationships between entities.
- Visualization of Ayurvedic Knowledge: Uses network maps and topic clusters to carry out structured analysis.
- Computational Integration of Traditional Medicine: Digitally saves and examines historical medical information.

The study emphasizes the efficiency of NLP in extracting and structuring Ayurvedic knowledge, making ancient medical texts more readable. Through the combination of NER, topic modeling, and knowledge graphs, it offers a scalable method for analyzing historical medical literature and digital preservation.

## 2. Related Studies

Natural Language Processing (NLP) has greatly contributed to electronic processing of ancient Indian manuscripts. Historical Sanskrit and Ayurvedic manuscripts hold huge volumes of information; however, being unstructured in nature, it is difficult to retrieve and interpret. The developments in NLP, machine learning(Lekshmi & Anoop, 2023), and constructing knowledge graphs have enabled extraction, classification, and organization of this information systematically. This section discusses significant research work consistent with our approach, namely auto-classification and knowledge extraction of the translated Sushruta Samhita.

Recent developments in natural language processing have dramatically enhanced the digitization and semantic structuring of archaic texts. Meltemi 7B was introduced by (Voukoutis et al., 2024), which is a large language model trained on a Greek text corpus comprising 40 billion tokens. This model optimizes text comprehension, automatic segmentation, and entity recognition so as to make historical documents accessible.

In the area of topic modeling(Anoop & Asharaf, 2020), neural embedding techniques have been very effective in analyzing historical texts. (Ginn & Hulden, 2024) applied dynamic topic modeling to BERT embeddings over the entirety of the surviving corpus of Roman literature(Rajeevan et al., 2025; Sharaf & Anoop, 2023). Their findings pitched neural models as providing more thorough qualitative insights and being less susceptible to fluctuations in hyperparameter settings than traditional statistical models like Latent Dirichlet Allocation (LDA). Besides this, BERTopic has been quite widely adopted for multilingual topic modeling(Anoop et al., 2016). (Mutsaddi et al., 2025) performed comparative studies with BERTopic, applied to Hindi short texts, establishing its superiority in terms of coherent topic generation against LDA(V.S et al., 2015). This makes it clear that transformer-based topic modeling has great potential in understanding textual themes across languages(Rajeevan, 2025).

Knowledge graphs have never been behind in arranging relations in medical and historical data. (Terdalkar & Bhattacharya, 2023) created an automated framework for building knowledge graphs from Sanskrit documents that enables querying-based retrieval. The system makes ancient texts accessible as formatted machine-readable information. The Āyurjñānam project also included the construction of an Ayurvedic text knowledge graph via manual annotation of drugs(Terdalkar et al., 2023). Such a proposition warrants great attention to evaluating more methods to catalogue and preserve traditional systems of knowledge in computable forms.

Natural language processing(Krishnan & Anoop, 2023) and knowledge graphs have really transformed the whole information retrieval system in Ayurveda. The Āyurjñānam program incorporated the knowledge graphs with user interfaces for effective querying and annotation (Terdalkar et al., 2023). Moreover, the Government of India has taken the initiative towards organizing and digitizing traditional medical knowledge through TDCL (TKDL, 2025). The TKDL uses standardized terminologies and classifications, making it mainstream, integrating itself in modern scientific discourse and legal perspectives to impede biopiracy.

The application of NLP, topic modeling(Anoop & Asharaf, 2021), and knowledge graphs has provided an empirical approach to studying ancient Indian medical manuscripts. Several studies have confirmed the use of semantic annotation, machine learning, and knowledge models for the extraction of relevant information from ancient manuscripts. The interoperability of computational linguistics and AI-based models makes access easier, allows better information extraction, and preserves traditional knowledge. Our research builds on these approaches, applying NER-based classification, BERTopic modeling, and knowledge graph construction to identify and examine dominant themes from the translated Sushruta Samhita, contributing to the wider digital humanities and historical text processing community.

## 3. Materials and Methods

The study uses NLP techniques along with Knowledge Graph approaches to extract, assess, and categorize data based on ancient Indian literature. We apply the Named Entity Recognition (NER), BERTopic-based topic modeling, and visualizing the networks with Neo4j to systematically categorize primary entities along with their interconnectivity from earlier medical and philosophical ideas.

### 3.1 Data Processing

The data is derived from NER-based entity extraction from Sushruta Samhita, an ancient Sanskrit text translated. The text was preprocessed in Python using prominent libraries such as spaCy (for NER), ScispaCy (for medical entity extraction), and NLTK (for preprocessing and tokenization of text). Extracted entities are concepts, scholars, treatments, diseases, plants, places, and classical texts, and are categorized for structured analysis. For topic modeling, BERTopic was utilized to identify salient topics within the text and Matplotlib and Seaborn were used for visualization of topic clusters. Preprocessed data was shaped into a Neo4j knowledge graph after extracted entity cleaning and validation using Cypher queries within Neo4j Desktop

A workflow diagram is shown in Figure 1 to detail the step-by-step process, from data extraction, topic modeling, entity classification, and building of a knowledge graph, to provide a transparent methodology for the analysis of ancient Indian knowledge systems.

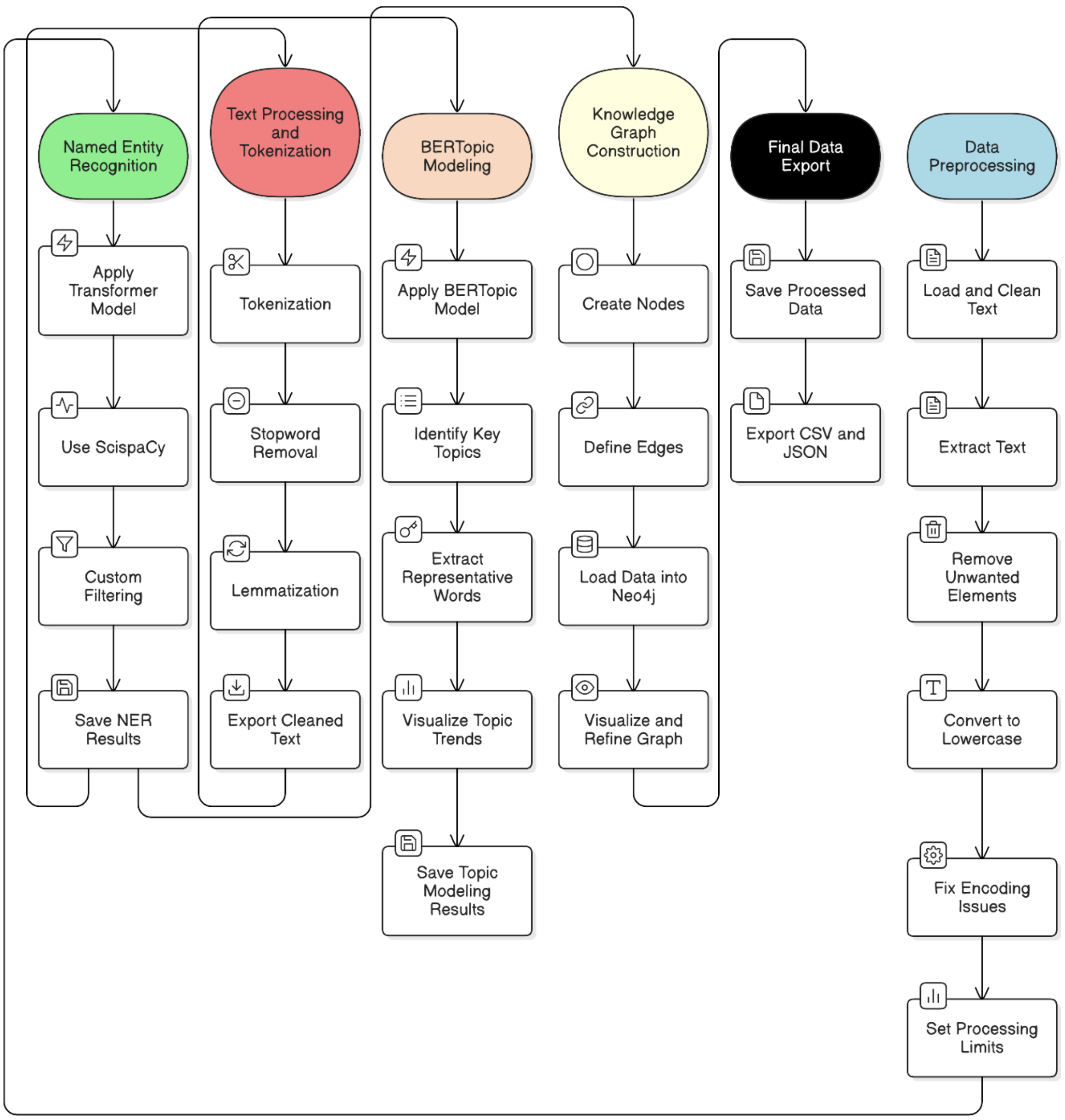

Figure 1: Workflow representation through flowchart

### 3.2 Named Entity Recognition (NER) for Knowledge Extraction

Named Entity Recognition (NER) is a critical task of Natural Language Processing (NLP). It is the task of recognizing and classifying significant entities such as disease, treatment, scholars, medicinal plants, and locations from text. In ancient Indian medical texts, NER assists in organizing knowledge. It makes it convenient to retrieve knowledge and relate significant concepts (Basha et al., 2023). Rule-based traditional approaches to NER are inefficient with the complexities of Sanskrit-English translation. Hence, machine learning and deep learning techniques are more efficient (Pandey & Nathani, 2024; Varghese & Anoop, 2022).

NER can be viewed as a sequence labeling task. Here, every word in a sequence is tagged with a label associated with its type of entity. For a given word sequence $X = \{x_1, x_2, ., x_n\}$, the task is to produce a sequence of labels $Y = \{y_1, y_2, ., y_n\}$. Here, $y_i$ is the type of entity (such as Disease, Treatment, Scholar).

One common approach is to use the Conditional Random Fields (CRF) model:

$$P(Y \mid X) = \frac{exp\left(\sum_t W^T f\left(y_{t,} y_{t-1}, X, t\right)\right)}{\sum_{Y'} exp\left(\sum_t W^T f\left(y'_{t,} y'_{t-1}, X, t\right)\right)}$$

where:

- f(y_t, y_{t-1}, X, t) represents feature functions that capture word-level information (e.g., POS tags, word embeddings).
- W is the weight vector learned during training.
- Y' represents all possible label sequences.

Alternatively, deep learning-based NER models, such as BiLSTM-CRF, employ bidirectional Long Short-Term Memory (LSTM) networks to learn the relationship between words. They also employ a CRF layer to produce structured predictions (Lample et al., 2016). The BiLSTM-CRF NER loss function is utilized as follows:

$$\mathcal{L} = -\sum_{i=1}^{N} \log P(Y_i \mid X_i)$$

where **N** is the number of training samples.

We employed ScispaCy, a dedicated NLP library trained on biomedical text, to identify entities corresponding to medical conditions, treatment, and plants. Additionally, we employed custom lexicons to augment the output by spelling variations and transliterations in the Sanskrit-English text. The identified entities were then sorted and mapped to Ayurvedic terms to enable structured representation for further analysis.

### 3.3 BERTopic Modeling for Thematic Analysis

BERTopic is a method of discovering topics in text using new methods known as transformer-based embeddings and clustering models. Unlike older methods such as Latent Dirichlet Allocation (LDA), BERTopic is able to detect distinct topic patterns by word meaning (Grootendorst, 2022). This approach is beneficial in the analysis of ancient Indian texts since they may be difficult to analyze due to complex language and transliterations (Blei et al., 2003; McInnes et al., 2020).

BERTopic combines three key components:

- Sentence Embeddings: Converts text into high-dimensional vectors using transformer models such as BERT, SBERT, or DistilBERT.

$$E(X) = f_{BERT}(X)$$

where X is the input text and E(X) is the embedding representation.

- Dimensionality Reduction (UMAP - Uniform Manifold Approximation and Projection): Reduces high-dimensional embeddings while preserving semantic relationships.

$$Z = UMAP\big(E(X)\big)$$

where Z is the reduced vector space representation.

- Clustering with HDBSCAN (Hierarchical Density-Based Spatial Clustering of Applications with Noise): Groups similar embeddings into topics based on density distribution.

$$T = HDBSCAN(Z)$$

where T represents topic clusters.

- Finally, Term Frequency-Inverse Document Frequency (c-TF-IDF) is applied to extract representative keywords for each topic:

$$c - TDF - IDF(t,T) = \frac{f_{t,T}}{\sum_{t' \in T} f_{t',T}} \log \frac{N}{\sum_{T' \in D} 1_{t \in T'}}$$

where:

- f (t, T) is the term frequency in a topic T.
- N is the total number of documents.
- 1_ {t \in T'} is an indicator function checking the presence of term t in topic T'.

We applied BERTopic to research the translation of the Sushruta Samhita, discovering meaningful topics regarding Ayurveda, medical procedures, and ancient texts. The model grouped similar words into distinct topics, allowing us to better understand ancient Indian medical knowledge (DhanushKumar, 2024;Grootendorst, 2024).

### 3.4 **Knowledge graph**

A knowledge graph is a structured representation of information, organized in nodes and edges, representing entities and relationships between them. This organized form allows interconnected data to be stored, retrieved and analyzed efficiently (Stegeman, 2024).

Formally, a knowledge graph can be represented as:

$$G = (V, E)$$

where $V$ is the set of nodes (entities) and $E$ is the set of edges (relationships).

Neo4j was used as the underlying database to implement and analyze the knowledge graph. Neo4j is a native graph database optimized for handling highly interconnected data. This allows for the efficient storage, querying, and visualization of graphs using Cypher, a declarative query language designed specifically for graph operations.

## 4. Results

This study employed certain techniques of Natural Language Processing (NLP) named entity recognition, BERTopic modeling, and network analysis to get structured information from the English translated version of the Sushruta Samhita. The results depicted in structured tables present the various aspects of our analysis. We further go on to discuss the results reported for each table.

### 4.1 Data Cleaning and preprocessing

This Table 1 summarizes the preprocessing steps performed on the extracted text from the Sushruta Samhita.

Table 1: Text Processing and data refinement

| Sl No. | Step | Description | Output Data |
|---|---|---|---|
| 1 | Text Extraction | Extracted raw text from the ancient translated PDF. | Extracted text with errors |
| 2 | Noise Removal | Removed page numbers, section headers, Roman numerals, and extra spaces. | Readable text without noise |
| 3 | Formatting Fixes | Fixed inconsistent formatting, spacing issues, and non-standard characters. | Formatted clean text |
| 4 | Named Entity Recognitio n (NER) | Identified key entities like diseases, treatments, scholars, and locations. | Categorized named entities |
| 5 | Topic Modeling (BERTopic ) | Extracted key topics and grouped related terms using BERTopic. | Topic clusters and themes |
| 6 | Final Data Preparation | Final cleaned structured datasets were prepared for knowledge graph integration. | Final processed datasets for analysis and visualization |

Sushruta Samhita analytical analysis and digital processing is structured to facilitate accurate entity recognition, topic modeling, and information extraction. The starting point is raw text extraction from the English-translated ancient text. The data has glitches such as inaccurately placed words, page numbers, and unusual formatting. Thus, unnecessary noise will be dismissed, such as headers tagged as redundant elements, Roman numerals, and excessive spacing leading to a better-organized text and easier reading. Besides this, formatting errors like variable spacing between characters and encoding errors were removed to give a well-accepted format for further analysis.

The utilized techniques involved NER in tagging and categorizing the essential entities such as diseases, therapies, researchers, medicinal plants, and locations. The extracted entities were then sorted prudently based upon their importance from the Ayurvedic medicine perspective. Following this, BERTopic modeling was introduced to extract latent thematic patterns from input text and cluster similar words under important topics such as disease categories, therapeutic options, and historic references. Finally, the findings were organized into the datasets which were then used to construct and visualize knowledge graphs thereby combining traditional medical knowledge with modern computational methods in a seamless way.

### 4.2 Named entity recognition (NER) and categorization

This Table 2 illustrates the classification of entities into various medical, historical, and textual categories. The Named Entity Recognition technique has been applied to recognize and categorize significant entities from the text. They have been, indeed, extremely helpful in the extraction of names, medical terms, scholars, diseases, treatments, plants, and place names.

Table 2: Categorization of key entities Identified in the Sushruta Samhita

| Sl. No | Category | Entities | Significance |
|---|---|---|---|
| 1 | Diseases | Kushtha, Prameha, Gulma, Vidradhi, Apachi, Jwara | These diseases are extensively discussed in classical Ayurveda texts. |
| 2 | Treatments | Sneha, Vasti, Dhuma, Triphala, Ghrita, Anutaila, Abhyanga | Ayurvedic treatment methods used for detoxification and healing. |
| 3 | Scholars | Susruta, Charaka, Dallana, Vagbhata, Jejjata, Chakrapani | Renowned scholars and contributors to ancient Indian medical literature. |
| 4 | Medicinal Plants | Haritaki, Nimba, Madhuka, Pippali, Amalaki, Guduchi, Ashwagandha | Common medicinal plants used in Ayurveda for healing and longevity. |
| 5 | Locations & Texts | India, Hastimsha, Rigveda, Sushruta Samhita, Ayurveda | Geographical and textual references providing historical context. |

Named Entity Recognition (NER) application has made it possible to systematically classify the important entities into their respective classes, leading to a more accurate conceptualization of the Ayurvedic medical framework. Classed as diseases, treatments, scholars, medicinal plants, and places/texts, these entities provide a clear thematic division between the medical and historical content in the text.

Diseases were categorized, including Kushtha (cutaneous diseases), Prameha (diabetes), Gulma (gastrointestinal diseases), Vidradhi (abscesses) and Jwara (fevers); other detailed descriptions can be found in Ayurvedic texts. Treatment procedures identified include, Sneha (oleation therapy), Vasti (medicated enemas) and Triphala (herbal preparation) that further underscored the procedures suggested in the text, essentially comprising detoxification and healing. Similarly, identification of great scholars like Susruta, Charaka, and Vagbhata highlighted the historical contribution of the medical pioneers towards the formation of traditional Indian medicine. Identification of herbal drugs like Haritaki, Nimba, and Ashwagandha highlighted the application of herbal medicine in Ayurveda, and identification of places and texts, i.e., Rigveda and the Sushruta Samhita itself, provided contextual depth to the study. These findings were then utilized in topic modeling and knowledge graph construction and bridged historic Ayurvedic knowledge into computerized modern-day analysis.

### 4.3 NER Entities by category

The Table 3 gives the frequency analysis of the Named Entity Recognition (NER) that classifies notable entities from the Sushruta Samhita. It highlights the frequency of local traditions (NORP), medical products (PRODUCT), and geographical locations (GPE, LOC), thus highlighting the historical, medical, and textual significance of the document.

Table 3: Distribution of named entity categories identified in the text

| Sl.No | Category | Entity Count |
|---|---|---|
| 1 | NORP | 565 |
| 2 | PRODUCT | 242 |
| 3 | DATE | 174 |
| 4 | CARDINAL | 109 |
| 5 | GPE | 79 |
| 6 | ORG | 74 |
| 7 | LOC | 64 |
| 8 | WORK_OF_ART | 58 |
| 9 | ORDINAL | 48 |
| 10 | QUANTITY | 39 |
| 11 | TIME | 30 |
| 12 | LAW | 23 |
| 13 | LANGUAGE | 4 |
| 14 | EVENT | 2 |
| 15 | MONEY | 1 |

The frequency of NER indicates the occurrence of NORP (565), referring to prominent local medical traditions, nationalities and religious or political groups found in Ayurvedic texts. PRODUCT (242) refers to medical preparations and instruments. Temporal entities like DATE (174), TIME (30), and ORDINAL (48) refer to the chronological information of treatments. GPE (79) and LOC (64) are utilized to denote geographical references, and WORK_OF_ART (58) is utilized to denote medical texts. This formal entity analysis proves useful in the acquisition of information regarding the historical, geographical, and medical context of the Sushruta Samhita.

The Figure 2 depicts the categorization of the recognized entities into broad interlinks of cultural associations, medicinal products, periods, numerical references, geographical locations and historical texts established through Named Entity Recognition (NER) applied to the Sushruta Samhita.

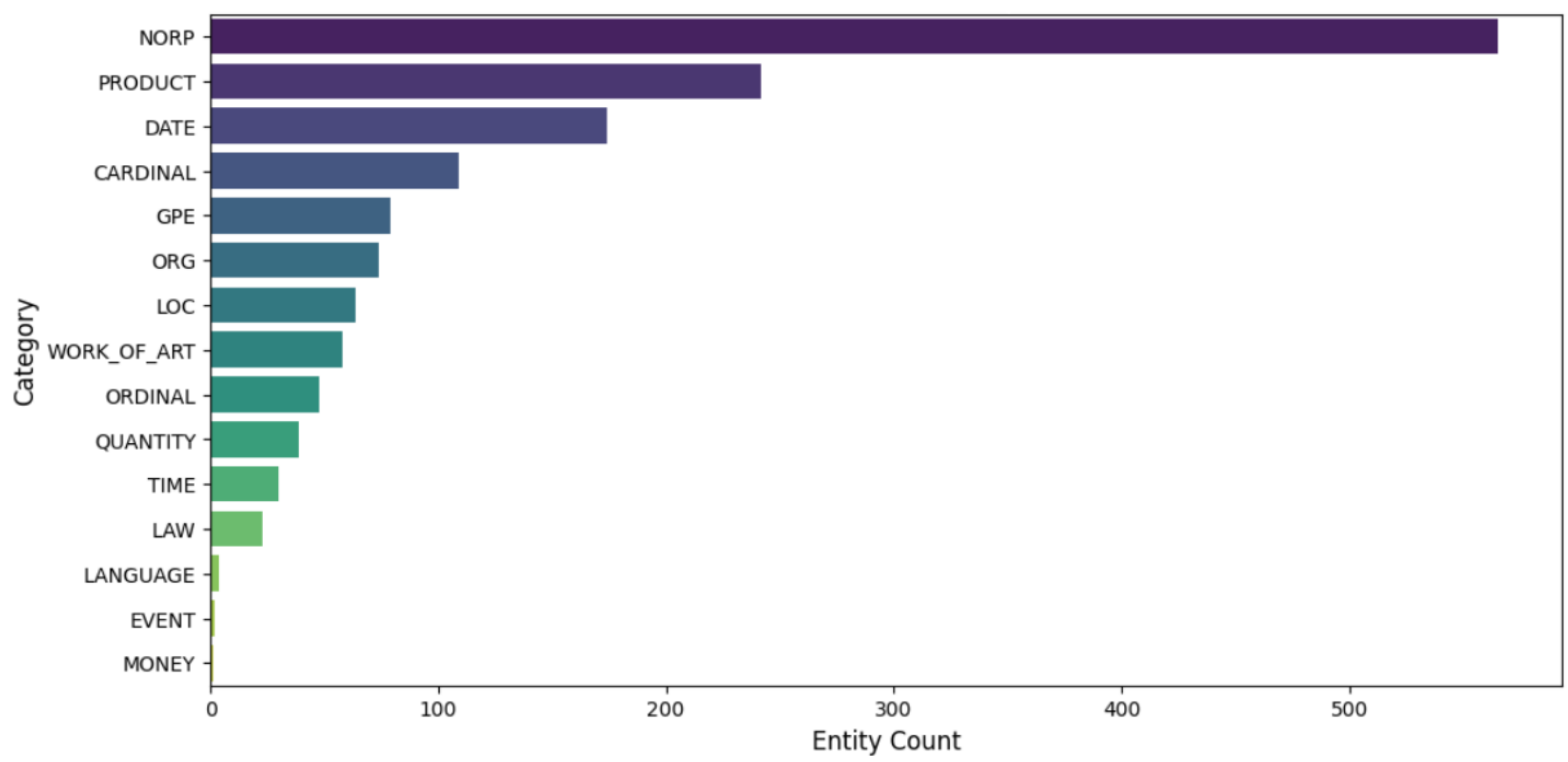

Figure 2: Distribution of extracted entities by category

The Figure 3 sunburst chart provides a hierarchical view of named entities in historical texts but particularly the Sushruta Samhita. It puts significant entities under headings such as Scholars, Diseases, Treatments, Medicinal Plants, and Locations/Texts.

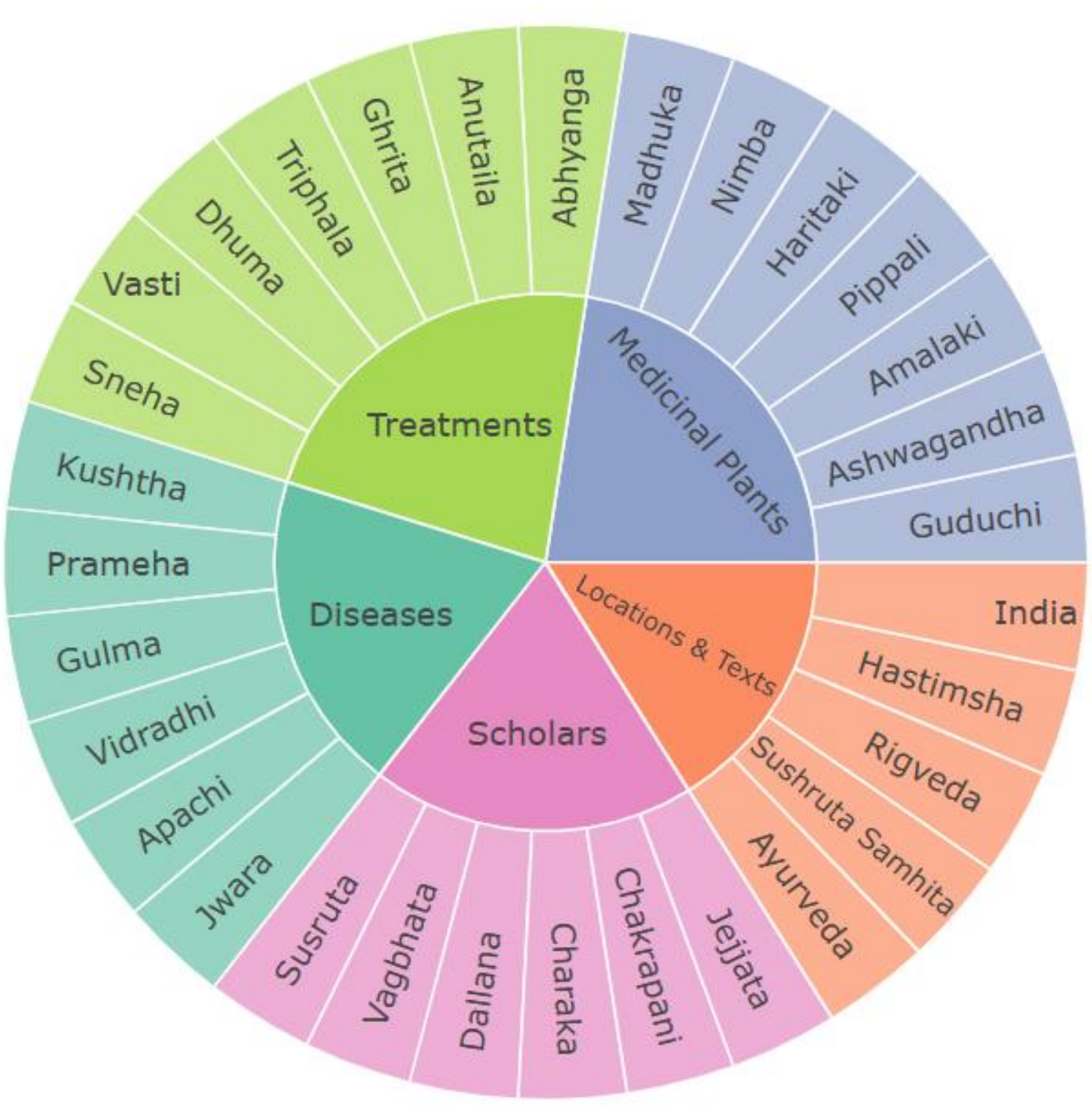

Figure 3: Hierarchical view of named entities in ancient texts

The class of "Scholars" contains great names such as Sushruta, Charaka, and Vagbhata to mark their contribution in the evolution of Ayurveda. The category of "Diseases" is formed considering different medical disorders, i.e., Kushta (cutaneous diseases), Prameha (diabetes), and Jwara (fever). The category of "Treatments" represents Ayurvedic treatments like Abhyanga (oil massage), Triphala (herbal mixture), and Ghrita (medicated ghee). The "Medicinal Plants" category indicates significant herbs such as Ashwagandha, Pippali, and Guduchi, indicating their medicinal use. Lastly, the "Locations & Texts" category tag major sources and places, such as Rigveda, Sushruta Samhita, and Ayurveda, thus providing information on texts and history.

The visualization provides an organized summary of the main concepts and entities in ancient Indian medicine texts.

**4.4 BERTopic modeling and topic extraction**

The Table 4 depicts the BERTopic modeling process employed for the Sushruta Samhita, ranging from preprocessing and tokenization through to topic extraction and clustering to finally arrive at an interpretation about an organized representation of core Ayurvedic principles, interventions, and historical contexts. It thus provides a processed topic dataset for visualization.

Table 4: Stepwise process of topic modeling using BERTopic

| Sl. No | Step | Description | Output Data |
|---|---|---|---|
| 1 | Preprocessing for Topic Modeling | Removed stopwords, numbers, symbols, and low-frequency words from the cleaned text. | Cleaned text without noise |
| 2 | Tokenization & Lemmatization | Tokenized text, applied lemmatization, and filtered out irrelevant terms. | Tokenized words with lemmatization |
| 3 | BERTopic Model Training | Trained BERTopic model on the processed text to extract key topics. | BERTopic topic clusters |
| 4 | Topic Extraction & Clustering | Clustered similar words into topics and assigned probability scores. | Topics with word distributions |
| 5 | Topic Interpretation & Labeling | Labeled each topic by analyzing top words and their context. | Refined and meaningful topic labels |
| 6 | Final Topic Data Preparation | Generated final structured dataset containing topics and their relationships. | Final topic-entity dataset for visualization and analysis |

Thematic patterns and interdependencies inherent in the Sushruta Samhita were identified using the modeling method of BERTopic. The pre-preprocessing performed and removed stopwords, digits, special characters, and rare words that were irrelevant noise. This was followed by Tokenization and Lemmatization, which decomposed the words into their roots while ignoring some unwanted irrelevant words. Through this, the data presentation was systematic, hence suitable for advanced topic modeling techniques.

Subsequently, the configuration of BERTopic was pretrained on drilled data, thereby learning important topics by discovering similar lexical units with clustering. Each topic was assigned probability scores to reflect how significant it was, how coherent it was with respect to other topics present in the data, and its significance. These topics were then cleaned and labeled such that the clusters encapsulated salient Ayurvedic principles, therapies, and historical references. Finally, a structured dataset was created that directly mapped the entities to the topics with a view toward enabling visualization, trend analysis, and knowledge graph integration to enable comprehensive examination of interdependencies within the text.

The Table 5 is the output of BERTopic modeling that identified the most significant themes that were found within the Sushruta Samhita. The model categorized similar terms to reveal key topics like medical cures, diseases, scholarly publications, medicinal plants, and historical/geographical sources. Each of the topics adds to the understanding of therapeutic practices, classification of disease, transmission of knowledge, herbal preparation, and the cultural background of Ayurveda, thus broadening understanding of the structure and content of this ancient medical text.

Table 5: Key topics identified in ancient medical texts using BERTopic

| Sl No. | Topic | Representative Terms | Theme | Importance |
|---|---|---|---|---|
| 1 | Medical Treatments | Vasti, Sneha, Triphala, Ghrita, Dhuma, Abhyanga | Ayurvedic therapies & purification techniques. | Explores the effectiveness of therapies used in ancient and modern Ayurveda. |
| 2 | Diseases & Disorders | Kushtha, Prameha, Vidradhi, Gulma, Jwara, Apachi | Common ailments & their Ayurvedic pathophysiology. | Helps in understanding how ancient medicine classified and treated diseases. |
| 3 | Scholarly Texts | Sushruta Samhita, Charaka, Vagbhata, Nidana, Chakrapani | Compilation of classical Indian medical texts. | Shows knowledge transmission through historical Ayurvedic scholars. |
| 4 | Medicinal Plants | Haritaki, Nimba, Madhuka, Amalaki, Pippali, Guduchi | Use of herbs & formulations in traditional medicine. | Highlights plant-based healing and pharmacological uses in Ayurveda. |
| 5 | Geographical & Historical | India, Hastimsha, Rigveda, Ayurveda, Ancient Texts | Cultural & geographical importance in Ayurveda. | Provides background on historical development and preservation of knowledge. |

The BERTopic model output illustrates significant thematic frameworks in the Sushruta Samhita, with indications of the medical, academic, and historical backgrounds of Ayurveda. The model was able to identify major topics such as Ayurvedic therapy, diseases & disorders, academic works, medicinal plants, and historical/geographical references. Every one of the topics was pinpointed by high-probability terms, revealing the dominant themes of the document. Accordingly, words such as Vasti, Sneha, Triphala, and Ghrita were linked with Medical Treatments to represent the over-arching status held by purification treatments and therapies in traditional Indian medicine. Disease categories such as Kushtha (skin diseases) and Prameha (diabetic-type illnesses) fell within the ambit of Diseases & Disorders since the intensive range of pathology under Ayurveda is huge.

Beyond the medical interventions, the findings also introduce the educational roots of Ayurveda, classifying classic texts and renowned authors like Sushruta, Charaka, and Vagbhata. The Medicinal Plants theme depicts herbal formulations applied in disease management, while Geographical & Historical includes citations to ancient texts, places, and cultural influences that have contributed to Ayurvedic knowledge. These findings not only offer proof of the structural coherence of the Sushruta Samhita but also of its ongoing utility in contemporary Ayurveda scholarship, positioning this topic modeling approach as a useful tool for classical Indian medical text analysis.

Figure 4 shows the topic word scores, with the most important words for each discovered topic. Every topic is expressed through its strongest words, illustrating prevailing ideas like medical treatments, diseases, academic references, medicinal plants, and historical background. This visualization is showing the most common words and aiding in finding central themes within Ayurvedic knowledge.

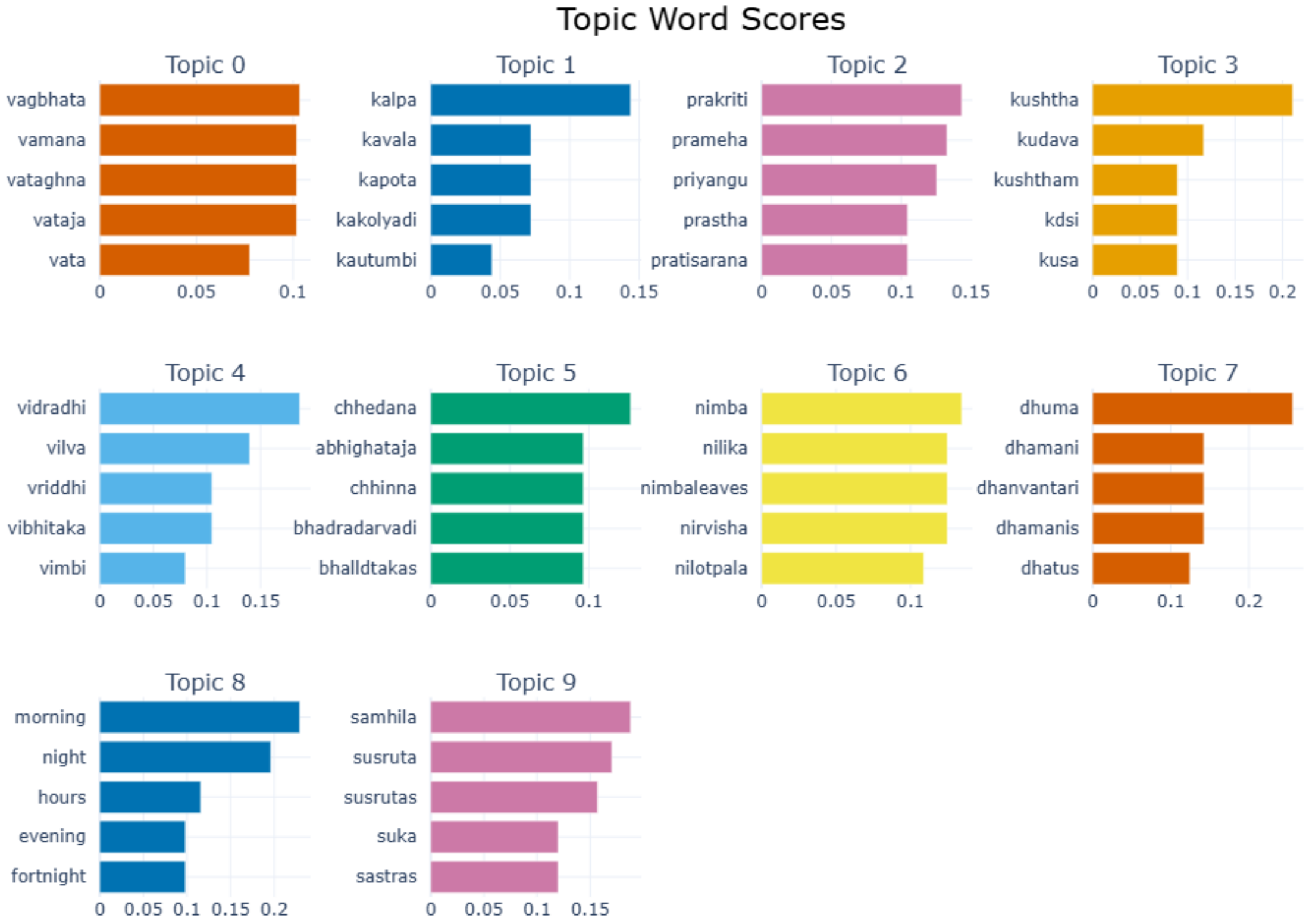

Figure 4: Topic word scores from BERTopic analysis

Figure 5 represents the top 10 most common topics present in the text. The most pervasive topics are "Medical Treatments" and "Diseases & Disorders," which indicate that the Sushruta Samhita is predominantly concerned with the practice of healing and disease management. The graph also indicates that surgical methods and herbal medicine have important roles in ancient Ayurvedic practice. The most common subjects further emphasize prominent Ayurvedic medical interventions and therapies, e.g., "vagbhata, vataja, vamana, vataghna," which refer to the contributions of Vagbhata, Vata disorders, and therapeutic emesis (Vamana). The term "Surgical Techniques" also encompasses words like "bhadradarvadi, chhinna, chedana" that are directly associated with surgical excision, wound management, and trauma management in ancient Indian medicine.

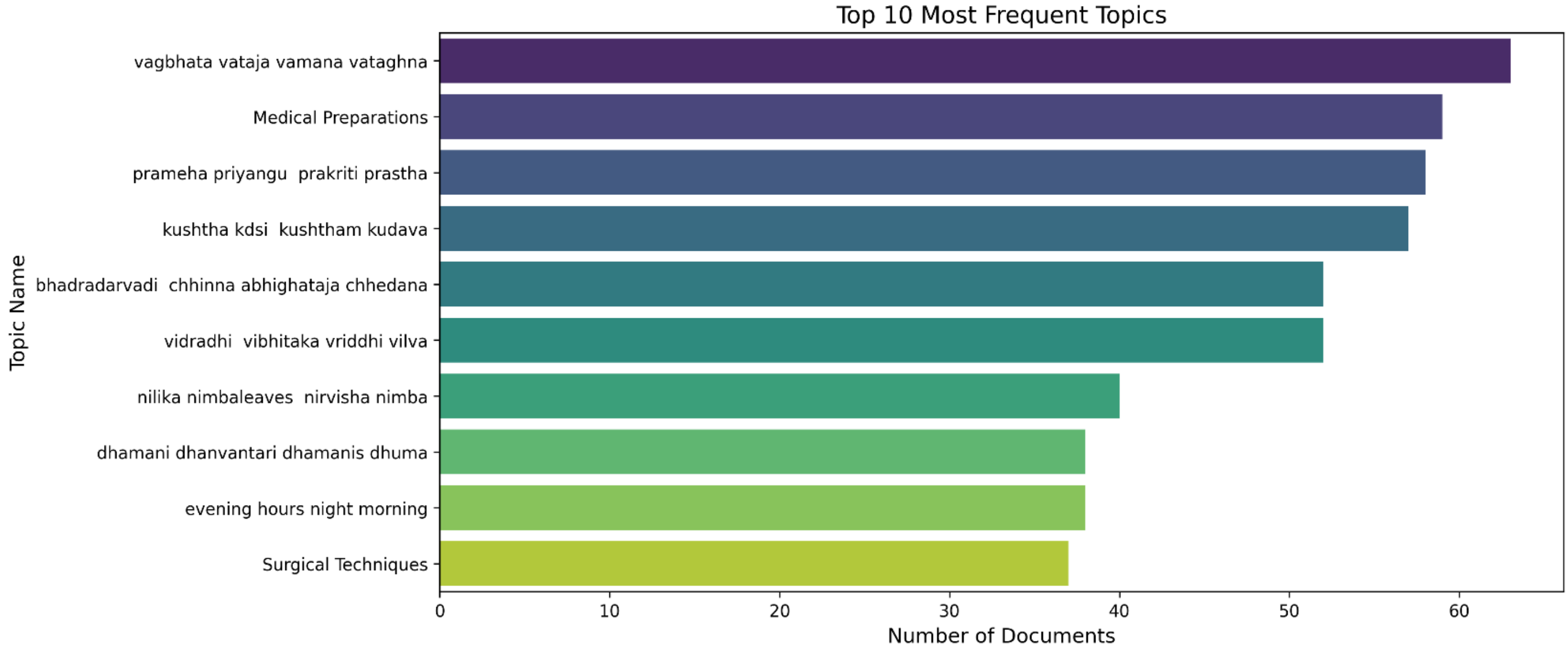

Figure 5: Top 10 most frequent topics identified in BERTopic analysis

A further important theme within Figure 5 is the stress placed on natural cures and herbal medicine. Reference to "nilika, nimbaleaves, nirvisha, nimba" signifies the importance placed on Neem (Nimba) and its antibacterial qualities. Finally, the use of "evening hours, night, morning" to note Ayurveda's concern for chronobiology and management of health and disease through circadian rhythms (Dinacharya) is relevant.

Figure 6 is an intertopic distance map, a two-dimensional representation illustrating the semantic relationship between topics. Dominant topics are represented by larger circles, and the closeness between them reflects conceptual similarity. The map illustrates closely packed topics in the center, reflecting overlapping themes in medical and procedural settings, while widely scattered topics reflect distinct thematic groups, e.g., particular diseases, surgical procedures, and herbal medicines. The size difference in circles indicates the prevalence of topics, with larger circles indicating more common topics. The clustering of central medical topics indicates fundamental discussions in the Sushruta Samhita, whereas the scattered topics indicate specialized subfields. These distributions can be used for knowledge graph construction and more precise topic labeling, resulting in a more structured representation of Ayurvedic medical knowledge.

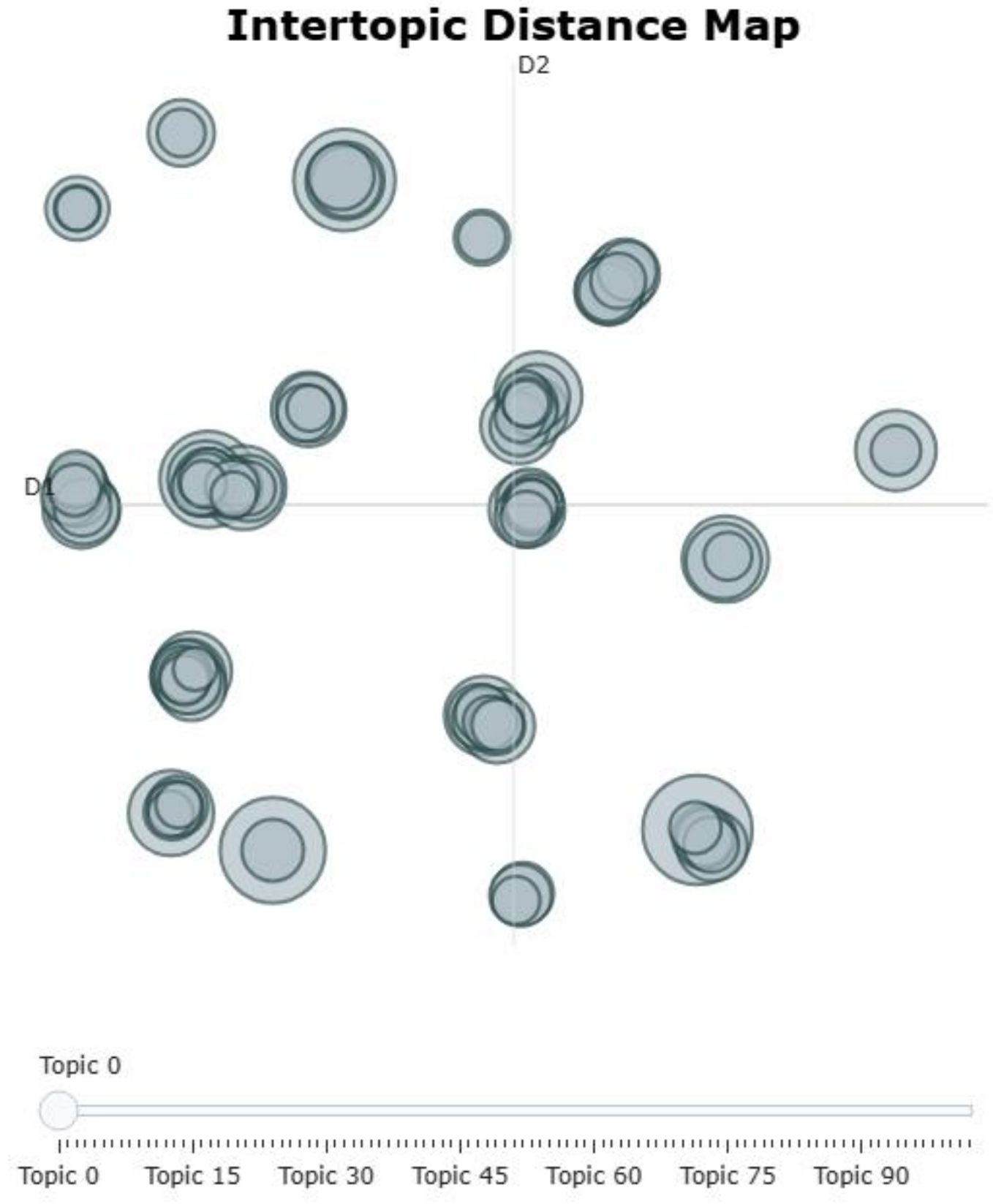

Figure 6: Intertopic distance map from BERTopic analysis

Figure 7 illustrates a 3D UMAP projection of BERTopic topics, which captures the high-dimensional structure of the dataset in a three-dimensional space. Points in the projection represent documents, and colors are used for various topic indices. The axes named UMAP Dimension 1, UMAP Dimension 2, and UMAP Dimension 3 give insights into the semantic distribution of topics. Points that are nearer each other represent higher thematic similarity, while points far apart represent more divergent topics. The blue-to-red color gradient separates various topic clusters and supplies clarity on the topic index and their relative positions in the dataset.

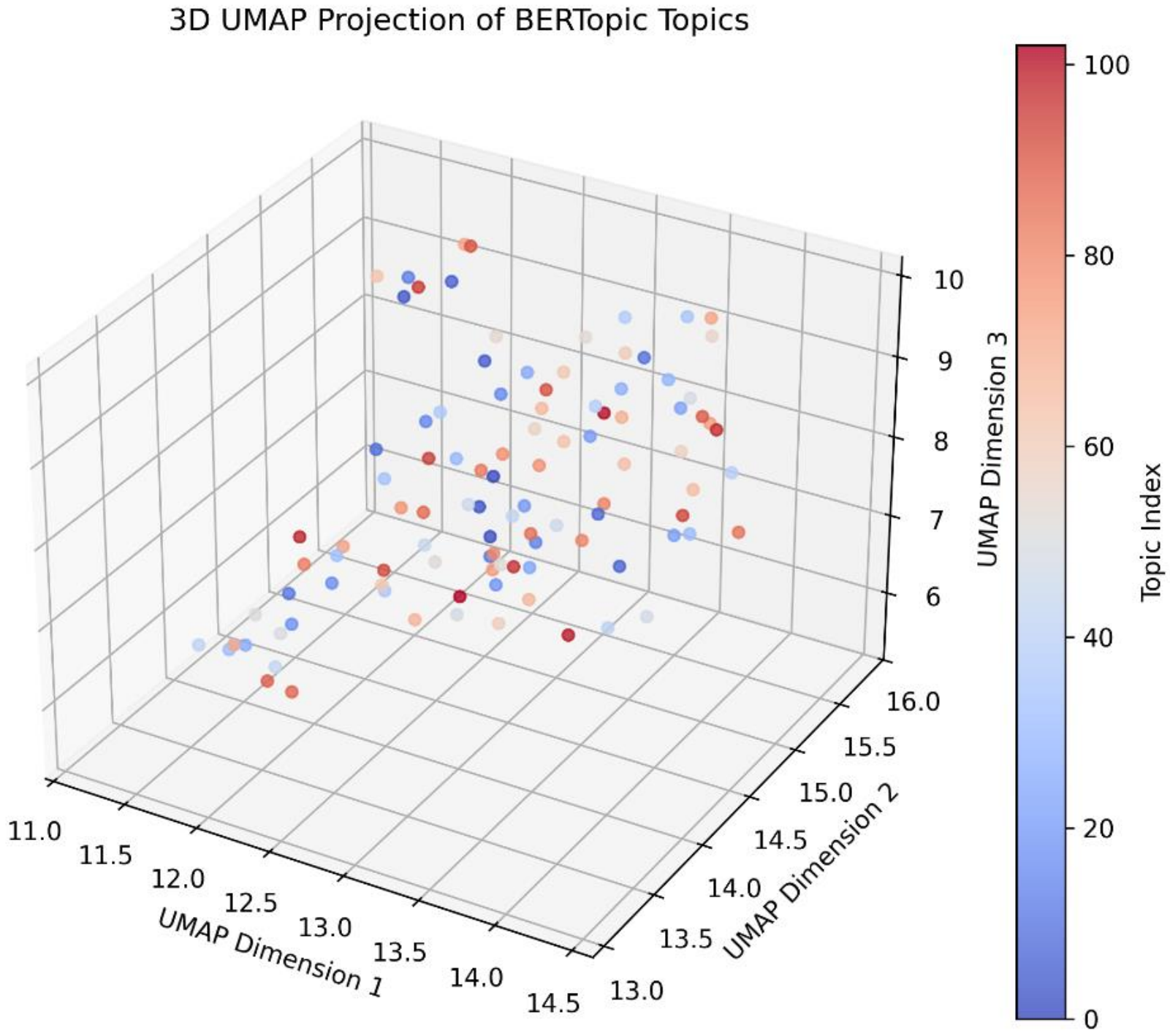

Figure 7: 3D UMAP projection of BERTopic topics.

The BERTopic model discloses the systematic method of the Sushruta Samhita, prioritizing medical therapy, disease classification, and drug knowledge. The discovered interrelationships between purification treatments, surgical proced ures, and herbal medicine underscore the holistic nature of Ayurveda. The fusion of topic frequency analysis and distribution mapping shows consistency in Ayurvedic discourse, whereas unique topic clusters signify in-depth areas of study. This computational method provides useful resources for organizing ancient medical knowledge to suit contemporary research. Future research can develop this analysis further through the inclusion of other Ayurvedic texts to facilitate comparative studies of medical principles through the ages.

**4.5 Knowledge graph representation and analysis**

Figure 8, the knowledge graph is a visual representation of relationships among major Ayurvedic entities such as classical works, experts, illnesses, therapies, and medicinal plants. 'Sushruta Samhita,' 'Charaka,' 'Triphala,' and 'Prameha' nodes are indicative of their connections to reveal insights on ancient Indian medicine.

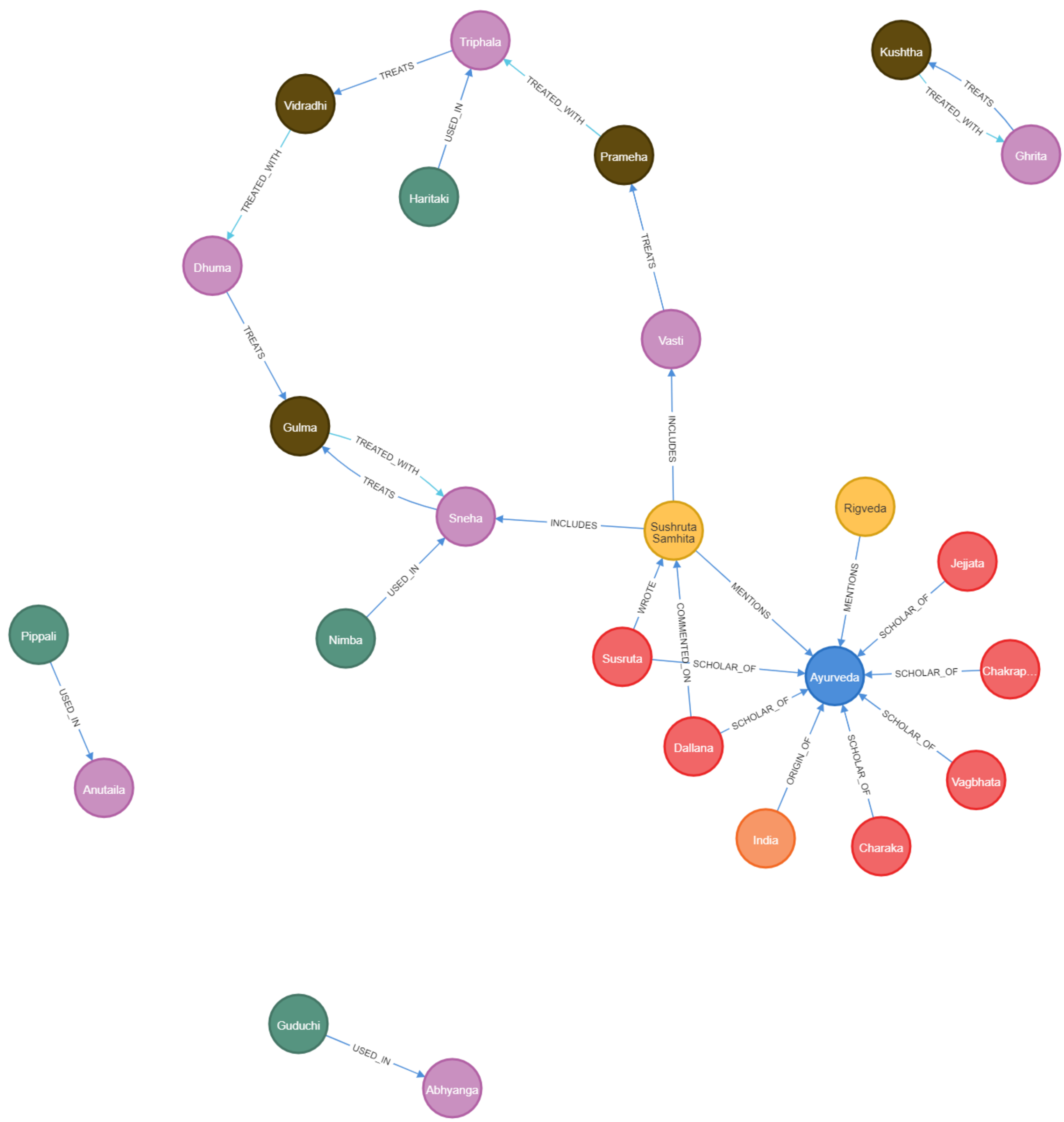

Figure 8: Knowledge graph representation of Ayurvedic concepts

The knowledge graph was built from Named Entity Recognition (NER) output and deployed in Neo4j Desktop with Cypher queries, projecting important entities from the Sushruta Samhita, a source Ayurvedic text. The graph contains 25 nodes and 26 relations, classifying entities into semantic roles. The nodes are color-coded to differentiate between different classes: concepts (blue) such as Ayurveda and Sushruta Samhita, treatments (violet) such as Sneha and Vasti, and diseases (brown) such as Vidradhi (abscess) and Prameha (diabetes). Plants (green) such as Nimba and Guduchi also denote the application of herbal medicines, and scholars (red) such as Susruta and Charaka are ancient scholars who worked on Ayurvedic knowledge. Locations (orange) such as India and Rigveda provide geographical and textual context.

The knowledge graph provides a comprehensive overview of the ancient Indian medical paradigm by structuring these entities and their interconnections, showcasing the relationship between diseases, treatments, plants, and scholars. In an easy-to-use repository styled document, the discoveries made therein will facilitate further mezzo-level research on progress, analysis, preservation, and retrieval of the Ayurvedic texts; the findings provide a glimpse into historical medical knowledge. The graph serves as a foundation for long-term Indology research with the help of natural language processing methods and could facilitate improvements such as semantic searches, knowledge extraction, or the possible integration of historical and current healthcare knowledge.

Knowledge graphs in ancient knowledge systems allow for the representation, retrieval, and analysis of ancient texts in a very structured way and help to preserve traditional wisdom in a digital format. Knowledge graphs provide the nodes of a graph with entities that may include but are not limited to concepts, scholars, treatments, diseases, plants, and places to allow for semantic search and enable the examination of complex relationships among components of ancient texts. They yet support text mining, entity linking, and cross-reference with present-day knowledge bases, which deepens the understanding of ancient practices yet enable machine learning and NLP-driven operations like automatic classification, reasoning, and verification of historical knowledge, making it feasible to integrate ancient knowledge with modern research. The process is highly applicable in various medical, philosophical, and literary traditions where ancient knowledge lends itself to contextual applications such as traditional medicine, linguistics, and cultural studies. Knowledge graphs make ancient texts more applicable, accessible, and relevant to researchers and practitioners alike through the creation of sound extraction of knowledge and contextual knowledge.

## 5. Recommendations.

The results of this research identify a number of areas that need to be improved and researched in the area of NLP-based classification and knowledge extraction from ancient Indian texts. One of the main issues that were identified was the misclassification of entities, which was due to the inadequacy of general-purpose Named Entity Recognition (NER) models when used with Sanskrit-derived terminologies and Ayurvedic jargon. To this end, future research should aim at the creation of NLP models specific to the domain that integrate personalized entity recognition and transliteration methods. This will contribute to improved accuracy in entity classification and a closer representation of Ayurvedic concepts.

In addition, although the BERTopic modeling method is efficient in the identification of thematic structures, it can be enhanced with the integration of ontology-based classification. A hybrid method that combines BERTopic with rule-based topic classification will enhance the readability of the extracted topics, thus reducing uncertainties in theme identification. Similarly, the improvement of the selection of coherence scores will enable better topic cluster identification, making the classification process more relevant.

The knowledge graph built using Neo4j provides a systematic representation of the identified entities. However, augmenting the knowledge graph with causal correlations between diseases, treatments, and medicinal plants will render it pragmatically effective. The integration of semantic reasoning and advanced query functionality will allow researchers to explore complex relationships in the Ayurvedic system. Moreover, interoperability with international medical ontologies will allow the integration of ancient and modern medical knowledge, thus promoting Ayurveda's adoption in modern healthcare research.

Finally, this research highlights the importance of digitizing and preserving ancient medical texts. Future studies should focus on building a standardized model for representing Ayurvedic knowledge, which would facilitate easy integration with historical, linguistic, and biomedical research. Interdisciplinary collaborations with linguists, medical experts, and computational scientists will further validate and enrich the derived knowledge, ensuring relevance and accuracy in interdisciplinary research.

## 6. Discussion and Conclusion

This research effectively applied Natural Language Processing (NLP) methods, i.e., BERTopic modeling and Knowledge Graphs (KGs), to label and extract structured knowledge from the English translation of the Sushruta Samhita, an ancient Sanskrit medical text. We could extract significant entities like diseases, treatments, researchers, and medicinal herbs using Named Entity Recognition (NER) so that we can have a structured understanding of the Ayurvedic medical system. Besides, BERTopic modeling revealed the underlying thematic structures in the text, linking semantically related terms and enabling an interpretable overview of fundamental medical concepts. The integration of these methods fostered a systematic approach to the analysis of historical medical texts, yielding new insights into Ayurvedic classification and its potential application in contemporary medical research.

One of the principal challenges was transliteration ambiguity and inconsistencies in the ancient text. Conventional NER models did not perform well with Sanskrit-derived words, which needed heavy custom preprocessing and rule-based fine-tuning for enhancing classification performance. Likewise, although BERTopic modeling was successful for topic clustering, tuning of topic labels was required to map coherently to Ayurvedic medical topics. These points identify the necessity for domain-specific NLP models learned on historical and Ayurvedic corpora, which would greatly enhance entity identification and thematic categorization.

The knowledge graph (KG) built based on the Neo4j was the central focus of the organization of relations among the extracted entities. Through connections among diseases, treatments, medicinal plants, and scholars, the KG offered a graph-based representation of Ayurvedic medical knowledge that supported semantic querying and inferencing. Utilization of graph databases in analyzing ancient texts has been found to be an efficient approach of structuring interlinked knowledge, thus offering a queryable and extensible platform for further research operations. The possibility of integration of such graphs with contemporary medical ontologies offers a strong potential for global incorporation of Ayurveda into contemporary healthcare systems.

In total, this research shows the application of NLP-driven classification and knowledge extraction in preserving and analyzing ancient Indian medical literature. The use of Named Entity Recognition (NER), topic modeling, and knowledge graph approaches gives a systematic and scalable solution to the organization of intricate historical knowledge. Future research needs to concentrate on the creation of domain-specific NLP models, improving the reasoning capability of knowledge graphs, and increasing the dataset to include a variety of historical texts. In doing so, this research contributes to the digitization and computational analysis of ancient medical knowledge, thus connecting traditional knowledge with data-driven approaches.